\documentclass[10pt,twocolumn,letterpaper]{article}

\usepackage[pagenumbers]{cvpr} %

\newcommand{\suppref}[1]{{#1}}

\usepackage{xspace}
\newcommand{\bench}{\textup{\textsc{AppCon-Bench}}\xspace}

\usepackage{graphicx}
\usepackage{amsmath}
\usepackage{amssymb}
\usepackage{booktabs}
\usepackage{enumitem}
\usepackage{cuted}
\usepackage{multirow}
\usepackage{xcolor}
\usepackage[table]{xcolor}
\usepackage[ruled,vlined]{algorithm2e}
\usepackage{tocloft}
\usepackage{titletoc}
\usepackage{svg}

\usepackage[normalem]{ulem}
\renewcommand{\paragraph}[1]{\vspace{.5em}\noindent\textbf{#1}}

\usepackage{tcolorbox}
\definecolor{cvprblue}{RGB}{0, 102, 170}
\colorlet{cvprblueBest}{cvprblue!30!white}
\colorlet{cvprblueSecond}{cvprblue!10!white}

\newcommand{\bestvlm}[1]{%
  \cellcolor{cvprblueBest}{\color{black} #1}%
}
\newcommand{\secondvlm}[1]{%
  \cellcolor{cvprblueSecond}{\color{black} #1}%
}

\DeclareRobustCommand{\bestlegend}[1]{%
  \begingroup
  \setlength{\fboxsep}{1.2pt}%
  \colorbox{cvprblueBest}{\color{black}\strut #1}%
  \endgroup
}
\DeclareRobustCommand{\secondlegend}[1]{%
  \begingroup
  \setlength{\fboxsep}{1.2pt}%
  \colorbox{cvprblueSecond}{\color{black}\strut #1}%
  \endgroup
}

\newcommand{\takeawaybox}[1]{%
    \begin{tcolorbox}[
        colback=cvprblue!6,
        colframe=cvprblue!55,
        arc=2mm,
        boxrule=0.5pt,
        left=3pt, right=3pt,
        top=2pt, bottom=2pt
    ]
    \textbf{Insights}

    #1
    \end{tcolorbox}%
}

\newcommand{\negative}[1]{%
  \textcolor{red!70!black}{#1}%
}

\newcommand{\scorechange}[2]{%
  #1\,\raisebox{-0.15ex}{\scriptsize #2}%
}

\definecolor{cvprblue}{rgb}{0.21,0.49,0.74}
\usepackage[pagebackref,breaklinks,colorlinks,allcolors=cvprblue]{hyperref}

\def\confName{CVPR}
\def\confYear{2027}

\title{\textsc{AppCon-Bench}: Perception and Misperception of \\ Material Appearance in Vision-Language Models}

\author{Santiago Jimenez-Navarro \quad Ana Serrano \quad Belen Masia\\
University of Zaragoza -- I3A\\
{\tt\small \{s.jimenez, anase, bmasia\}@unizar.es}
}

\begin{document}

\twocolumn[{%
\renewcommand\twocolumn[1][]{#1}%
\maketitle
}]

\begin{abstract}

Identifying the same material across changes in illumination or geometry is a critical ability for robust vision systems. Humans exhibit remarkable material appearance constancy, being able to recognize the intrinsic physical properties of an object's surface material across variations in extrinsic, confounding factors such as the lighting or the shape of such object. Vision-Language Models (VLMs) demonstrate strong performance across a wide range of semantic vision tasks, yet their ability to interpret subtle mid-level perceptual attributes remains less understood. We introduce \bench, a benchmark comprising five tasks designed to assess material appearance constancy under controlled variations in geometry and illumination, and evaluate representative VLMs alongside human observers, investigating their ability to maintain consistent material judgments across viewing conditions and experimental contexts. After thorough analysis, we identify several key findings that reveal both similarities and systematic differences between human and VLM material appearance perception, highlighting the value of evaluating perceptual successes and failures beyond overall accuracy and informing areas of improvement for current VLMs.

\end{abstract}

\section{Introduction}
\label{sec:intro}

Large Vision-Language Models (VLMs) have shown strong capabilities across a wide range of visual tasks, from image captioning~\cite{hua2025finecaption, lu2025benchmarking} to specialized applications such as metric depth estimation~\cite{yeh2026depthlm} and white balancing~\cite{li2026vlmcc}. However, strong performance on downstream tasks does not necessarily imply human-like visual perception. While recent interpretability methods investigate internal attention mechanisms~\cite{kaduri2025s, liu2026aligning} and learned visual representations~\cite{ferrando2026language}, understanding how VLM perceptual behavior compares with human judgments remains an open challenge.

\begin{figure}[t]
  \centering
   \includegraphics[width=\linewidth]{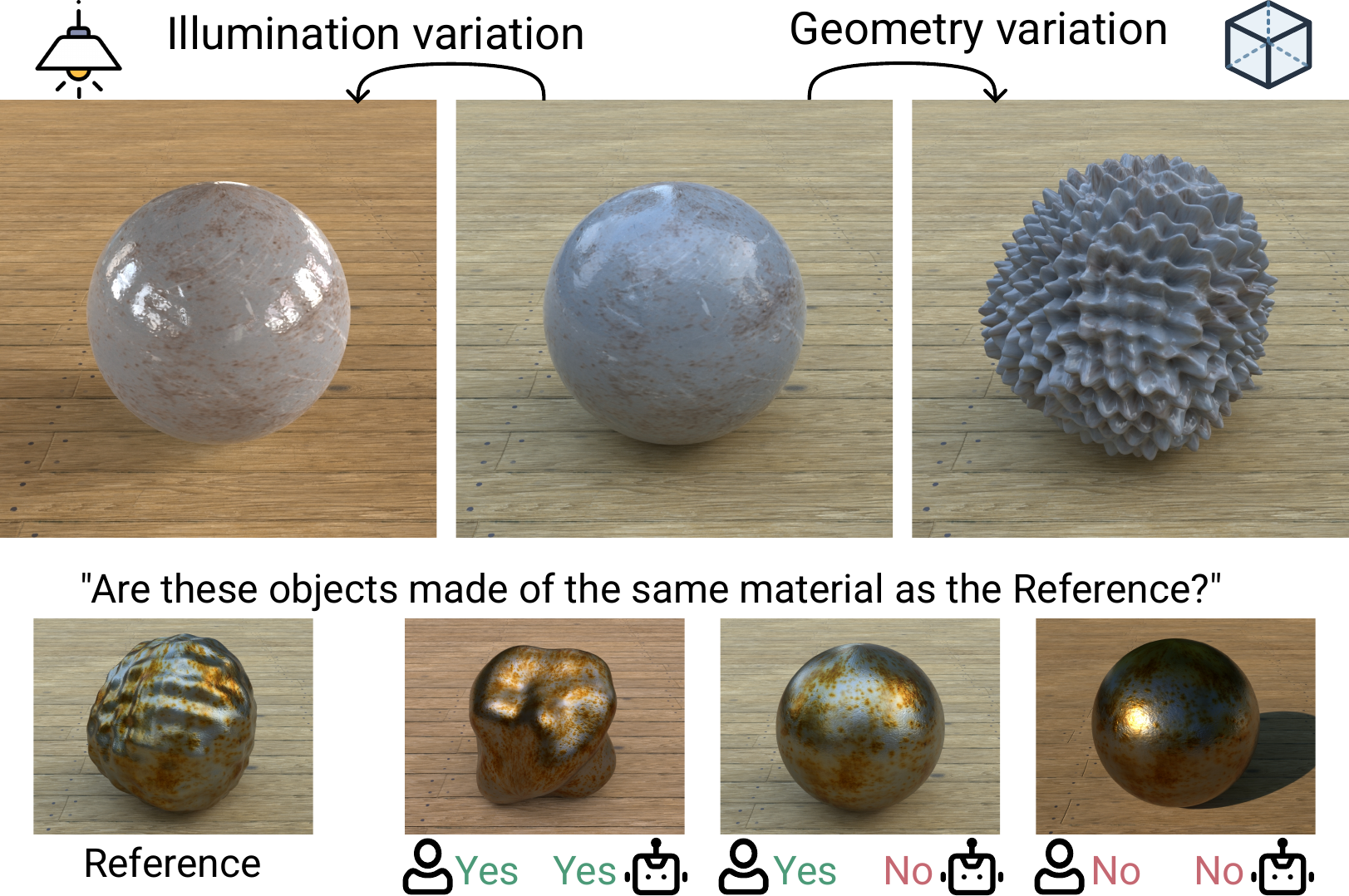}
   \caption{
   \textbf{Top:} Three objects made of the same physical material exhibit different appearance under changes in illumination and geometry.
   \textbf{Bottom:} Although all four objects share the same physical material, humans and VLMs may differ in their ability to recognize them as such, exhibiting distinct perceptual constancy.
   }
   \label{fig:1_constancy}
\end{figure}

Material appearance provides a particularly challenging setting for this investigation. The appearance of an object arises from complex interactions between surface reflectance, geometry, illumination, and viewpoint~\cite{fleming2017material, fleming2014visual, lagunas2021joint}, making the inference of intrinsic material properties inherently ambiguous (\Cref{fig:1_constancy}, top). Nevertheless, the human visual system can maintain stable material judgments despite changes in viewing conditions, an ability known as \emph{perceptual constancy}. Importantly, such constancy is not always preserved, and observers may systematically misperceive physically identical materials (\Cref{fig:1_constancy}, bottom).

VLMs' strong Visual Question Answering (VQA) capabilities enable the adaptation of psychophysical experiments to probe their sensitivity to visual attributes. Recent studies have investigated VLM perception of color~\cite{liang2025colorbench} and surface slant~\cite{zhang2026anchored}, as well as color, size, and shape constancy~\cite{sun2025probing}. However, comparatively little is known about VLM constancy for 
material appearance attributes such as
glossiness and metallicness. In particular, we ask whether VLMs maintain material judgments across changes in viewing conditions, and whether their perceptual successes, failures, and ambiguities resemble those of humans. 
To investigate these questions, we present \bench, a benchmark comprising five psychophysics-inspired tasks designed to study perception and misperception of material appearance in VLMs (\Cref{sec:experiments}). We carefully design 3,000 trials and collect 21,000 human responses, alongside an equal number of responses from each of the 12 VLMs tested (\Cref{sec:experiments:setup}). We confirm the validity of the benchmark by obtaining results with human participants that align with previous studies on vision science (\Cref{sec:results}). By comparing human and VLM judgments across different %
tasks, we investigate not only their overall performance but also their perceptual biases and failure patterns. Our analysis reveals several key findings:

\begin{itemize}
    \item humans and VLMs differ on their robustness to illumination and geometry changes;
    \item VLMs, like humans, are able to leverage contextual information to improve material recognition under challenging conditions;
    \item when the number of correct responses is not fixed, VLMs exhibit a systematic over-selection bias, revealing response strategies that differ substantially from those of humans;
    \item humans and VLMs show substantial agreement in perceptual difficulty; and
    \item the best-performing VLMs reproduce human-like illumination-dependent trends in perceived glossiness, but differ from humans in their sensitivity to scene-induced appearance changes.
\end{itemize}

These findings contribute to a better understanding of material perception in VLMs, providing insights into their capabilities and limitations. To support further research, we will publicly release \bench, including the stimuli, questioning protocols, and collected human and VLM responses.

\section{Related work}
\label{sec:rw}

\paragraph{Material appearance perception.}
Visual information is central to how we perceive objects and interact with our surroundings.
In this context, material appearance results from the complex interplay between surface reflectance, object shape, lighting conditions, and viewpoint, which makes inferring intrinsic material properties from visual appearance inherently ambiguous~\cite{fleming2003real, fleming2017material, fleming2014visual, lagunas2021joint, xiao2025material}.
Despite this ambiguity, the human visual system can preserve stable judgments of intrinsic material properties across varying viewing conditions~\cite{fleming2017material}, an ability known as \emph{perceptual constancy}.
While color constancy has been extensively studied~\cite{troost1991naming, foster2011color, radonjic2015color, bannert2017invariance, brainard2018naturalistic}, 
 middle-level perceptual attributes central to material appearance, like glossiness or metallicness, have received comparatively less attention in the literature on perceptual constancy. 
Glossiness, which describes the perceived shininess or luster of a surface~\cite{chadwick2015perception, morimoto2026human, bousseau2013gloss, storrs2021unsupervised, prokott2021gloss, guerrero2024predicting}, has been studied by 
Morimoto \etal~\cite{morimoto2023color}, who reported limited gloss constancy across variations in illumination and geometry, with highly consistent perceptual errors across observers. 
Similarly, Wendt \etal~\cite{wendt2010disparity} found that providing additional cues about color, disparity, or motion can help improve gloss constancy,  
while Olkkonen \etal~\cite{olkkonen2011joint} found strong interactions between geometry and illumination in gloss perception that could not be explained by luminance histogram statistics. 
Regarding metallicness, which relates to the extent to which a surface appears metallic~\cite{harvey2021low, todd2018visual}, Norman \etal~\cite{norman2020effects} showed that changes in illumination can cause metallic surfaces to be perceived as nonmetallic, and vice versa.
It is worth noting that beyond perceptual accuracy, reproducing systematic human errors can provide valuable evidence of behavioral alignment~\cite{storrs2021unsupervised, morimoto2026human, mao2025evaluating, zhang2026anchored}.

\paragraph{Appearance perception in VLMs.}
Vision-Language Models (VLMs) enable multimodal reasoning by combining a textual decoder with a vision encoder through a learned adapter, allowing visual and textual tokens to interact with each other~\cite{alayrac2022flamingo, li2023blip, liu2024improved, bai2023qwenvl}.
While VLMs perform strongly on semantic-related vision tasks like visual grounding or image captioning~\cite{lu2025benchmarking, kang2025your, liu2024mmbench, yu2023mmvet}, recent works show that their performance in low- and mid-level visual features may be limited~\cite{tangtartharakul2026visual, tortei2026visres}. 
VLMs' strong Visual Question Answering (VQA) capabilities enable the adaptation of traditional psychophysical experiments to probe their sensitivity to visual attributes, although behavioral responses alone cannot reveal how these attributes are internally represented. Recent studies have employed this approach to investigate color perception~\cite{liang2025colorbench}, visual illusions~\cite{mao2025evaluating, sun2026vlms}, or surface slant~\cite{zhang2026anchored}.
Their findings reveal both human-like perceptual biases and discrepancies with human perception, sometimes reflecting a reliance on learned priors rather than visual evidence~\cite{sun2026vlms, luo2024probing}. Closer to our work, Sun \etal~\cite{sun2025probing} investigate whether VLMs exhibit color, size, and shape \emph{constancy}. In this paper, we are the first to look into whether VLMs exhibit \emph{material appearance constancy}, with a focus on glossiness and metallicness, and compare both their successful judgments and systematic perceptual failures with human observers.

\section{\bench}
\label{sec:experiments}

We introduce \bench, a benchmark designed to evaluate material appearance constancy, with a focus on glossiness and metallicness, in humans and VLMs. Inspired by traditional vision science studies~\cite{brainard2018naturalistic, fleming2003real, olkkonen2010perceived}, it combines naturalistic stimuli with tasks presented through psychophysical experiments for humans, and Visual Question Answering (VQA) for VLMs.

\begin{figure*}[ht]
  \centering
   \includegraphics[width=0.96\textwidth]{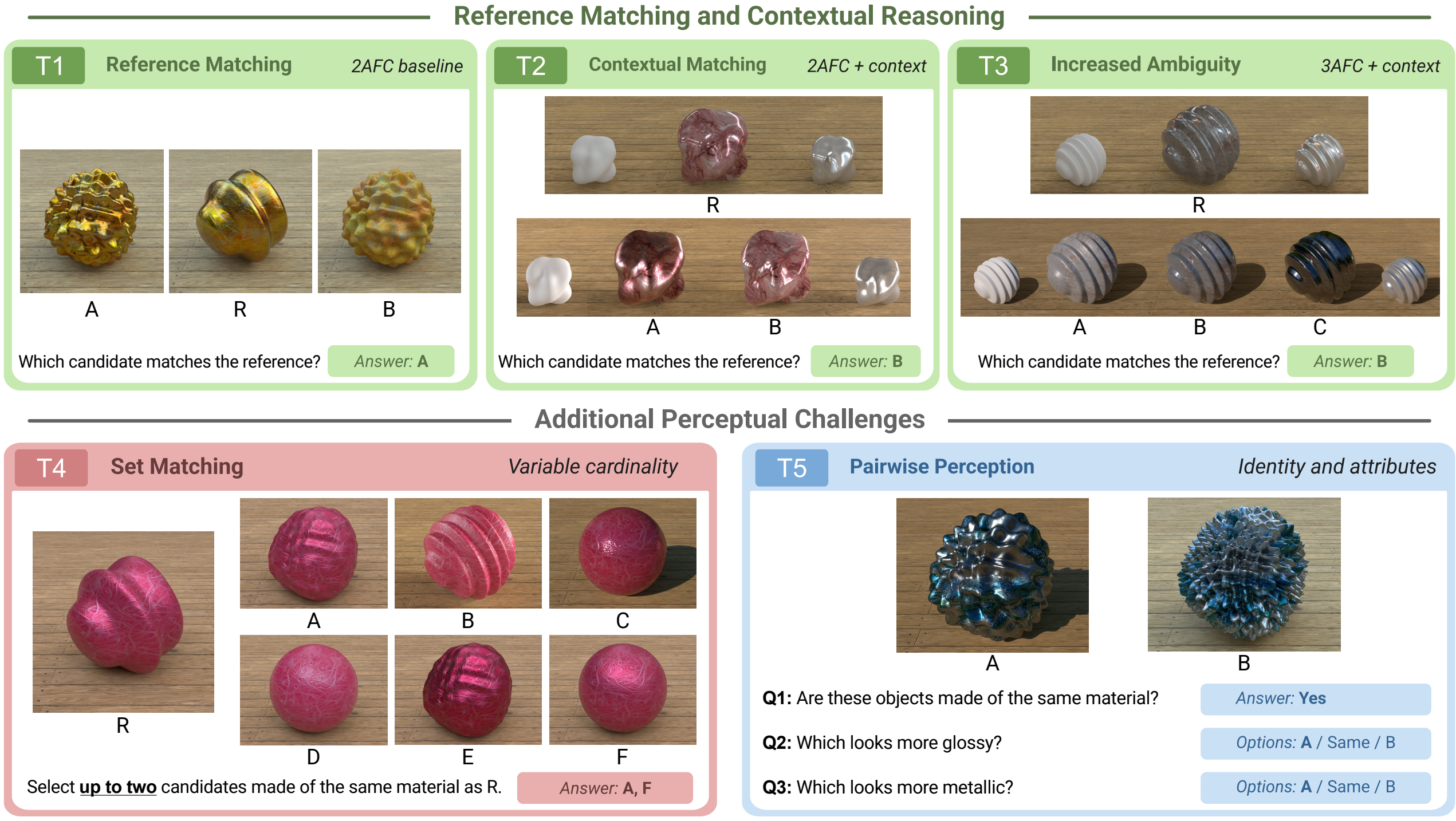}
   \caption{
   \textbf{Overview of the five tasks in \bench}. We show representative trials with their corresponding questions and ground-truth answers. Comprising 3,000 trials, \bench enables systematic evaluation of material appearance perception and constancy in humans and VLMs. Note that T5:Q2 and T5:Q3 are purely perceptual questions which have no objective ground truth.
   }
   \label{fig:benchmark_overview}
\end{figure*}

\subsection{Stimuli}
\label{sec:experiments:stimuli}

\begin{figure}[t]
  \centering
   \includegraphics[width=\linewidth]{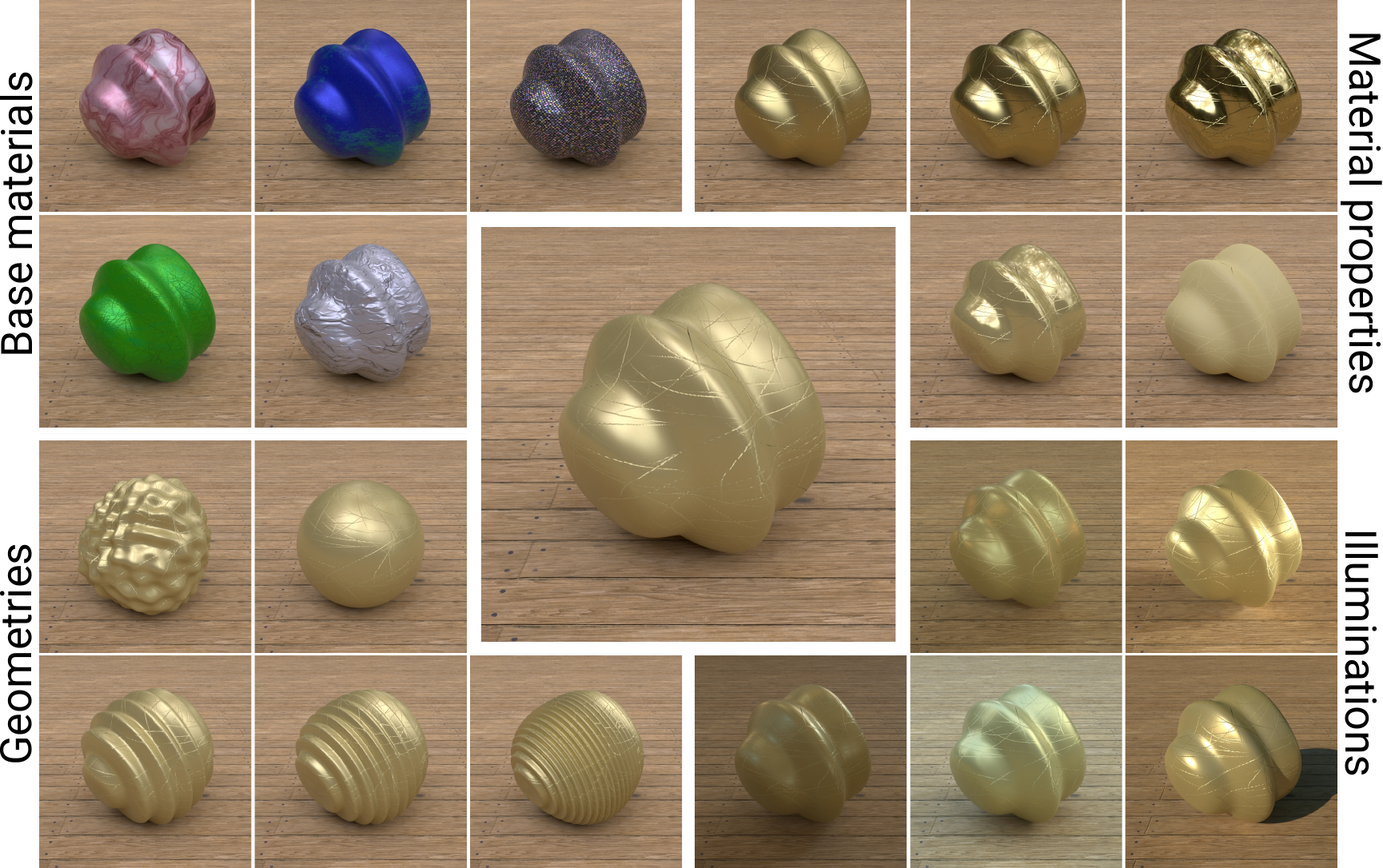}
   \caption{\textbf{Generative factors} used to render the photorealistic stimuli shown in the experiments. Starting from the central reference image, we illustrate variations in \textit{base materials}, \textit{material properties}, \textit{geometries}, and \textit{illuminations}.
   }
   \label{fig:3_generative}
\end{figure}

Our stimuli consist of images of single objects, with varied materials, shapes and illuminations. We opt for realistic visual stimuli to evaluate material appearance perception under representative viewing conditions, since deviations from natural image statistics and appearance cues can influence human perceptual judgments~\cite{wiebel2015statistical} and VLM responses~\cite{li2024mvp}. 
Since real-world photographs offer limited control over geometry, illumination, and material properties, we use the physically-based Mitsuba 3 renderer~\cite{Mitsuba3} to generate photorealistic stimuli with controlled scene parameters. %
To provide a naturalistic setup, we render individual objects made of different materials, placed atop a wooden surface.
We systematically vary the material of objects, their shape (geometry), and the illumination of the scene (\Cref{fig:3_generative}).

\paragraph{Materials.} 
We select fifteen spatially-varying materials from the \textit{MatSynth} dataset~\cite{vecchio2024matsynth}, comprising high-quality physically-based rendering (PBR) materials represented by Spatially-Varying Bidirectional Reflectance Distribution Functions (SVBRDFs). This representation allows us to modify reflectance-related parameters while preserving realistic material appearance. For each base material, we combine three variations of the \textit{Roughness} and \textit{Metallic} maps, yielding nine configurations of material properties per source SVBRDF, resulting in a total of 135 materials (see \suppref{supplementary material} for more details). 
Note that these modifications provide controlled variations in material parameters, yet do not necessarily produce perceptually-linear changes in appearance.

\paragraph{Geometries.} 
We select ten different shapes or geometries, with varying surface curvature and relief, including a \textit{sphere}, a \textit{blob}~\cite{vangorp2007influence}, and sphere variants with \textit{ridges} or \textit{bumps} at four spatial frequencies each. These variations allow us to investigate how surface geometry affects material perception~\cite{storrs2021unsupervised, marlow2012perception}. %

\paragraph{Illuminations.}
We use six high-dynamic-range (HDR) environment maps from \textit{Poly Haven} (\url{https://polyhaven.com/hdris}) selected to produce diverse illumination conditions. We provide details of their selection in the \suppref{supplementary material}.

\subsection{Tasks}
\label{sec:experiments:experiments}
We design five different tasks, leveraging the stimuli introduced in  \Cref{sec:experiments:stimuli}, to assess material appearance perception in humans and VLMs. We describe them here, while an example trial of each is shown in \Cref{fig:benchmark_overview}.

\paragraph{T1 -- 2AFC reference matching. } Given a reference and two candidates, the task is to choose which candidate depicts the same material as the reference (two-alternative forced choice, 2AFC). In each trial, both candidates share the same geometry and illumination, and either geometry or illumination varies between the reference and the candidates (this is the case in tasks T2 to T4 as well). This task allows assessing whether the same material can be recognised when presented with a different geometry or illumination, i.e., whether the human or VLM exhibit material constancy.

\paragraph{T2 -- 2AFC with added context. } The setup is like that of T1, but introducing additional contextual probes alongside the reference and candidates, allowing us to investigate whether contextual information improves material appearance constancy~\cite{brainard2018naturalistic} for humans or VLMs.

\paragraph{T3 -- 3AFC with added context. } The setup further extends T2, introducing a third candidate (different from the other two), allowing us to examine the effects of additional distractors on matching performance of humans and VLMs. 

\paragraph{T4 -- Multiple candidates and variable answer cardinality. } In each trial, given a reference, the task is to identify all material matches to the reference among six candidates, of which up to two are correct. Unlike previous ones, this task allows a variable number of responses, enabling the analysis of false positives and negatives under increased task complexity and the comparison of human and VLM behavior under increased uncertainty. %

\paragraph{T5 -- Pairwise comparisons: material identity and perceived attributes. } Each trial features a pair of images (no reference), and three questions are sequentially asked: (Q1) whether the two images depict objects made of the same material, (Q2) which object appears more glossy, and (Q3) which object appears more metallic. Unlike previous tasks, here multiple scene factors (geometry, illumination and material) may vary simultaneously. Note that two objects may be made of the same material, yet one appear glossier than the other due to the confounding factors of illumination and geometry. This task allows us to examine material identity judgments, as well as their relationship with perceived glossiness and metallicness, potentially identifying factors associated with successful and unsuccessful constancy. %

\subsection{Experimental setup }
\label{sec:experiments:setup}

\paragraph{Trial configuration.} When building the different trials in a task, rather than exhaustively testing all stimuli combinations (many would not be very informative, e.g., 2AFC where one of the candidates features a very different material from the reference), we curate them based on their level of difficulty, estimated by a combination of image statistics and editing distances (see \suppref{supplementary material} for additional details and validation of this estimation). Importantly, tasks T1 to T3
use the same setups (combination of reference and candidates) to allow for a controlled analysis of the effects of adding contextual information (T2) and using an additional distractor (T3; here, another candidate of the same estimated level of difficulty is added). 
T4 further extends the setups used in T3 by introducing the necessary matches and confounders to create a six-option setup.

\paragraph{Human responses.}
We collected human judgments through the \textit{Prolific} platform for each benchmark task (\Cref{sec:experiments:experiments}), with a total of 420 participants (84 per task). To obtain robust responses and assess human agreement, seven independent responses per trial were collected. Further details on data collection and participant demographics are provided in the \suppref{supplementary material}.

\paragraph{VLM inference data.}
We evaluated 12 open-weight VLMs from the Qwen~\cite{bai2025qwen3}, Gemma~\cite{team2026gemma}, LLaVA~\cite{liu2024llavanext}, and InternVL~\cite{wang2025internvl3} families, spanning multiple model sizes. Each trial was evaluated seven times per model using different random seeds to assess response variability. Model configurations and inference details are provided in the  \suppref{supplementary material}.

\section{Results}
\label{sec:results}

\begin{table*}[t]
  \centering
  \caption{
    Overall performance and VLM agreement with human observers. For each task, Accuracy denotes the proportion of correct responses, and Cohen's $\kappa$ is used to compute the agreement of each VLM with humans. McNemar's test is used to test whether the difference in accuracy between a VLM and the humans is statistically significant: all of them are except those marked with $^\dagger$. For T4, multi-label selection performance is summarized by the mean sample-level F1 score and the candidate-level Matthews correlation coefficient (MCC), computed across the six candidate options.
    Also, the setup-level Spearman correlation $\rho_{\mathrm{H}}$ and Kendall's $\tau_{\mathrm{H}}$ measure agreement with humans in the relative difficulty of the setups.
    Regarding T5, performance measures correspond to Q1, for which ground truth is available. 
    \bestlegend{Blue} and \secondlegend{light blue} cells indicate the best and second-best VLM point estimates in each metric.
  }
  \label{tab:4_overall_model_performance}
  \begingroup
  \scriptsize
  \setlength{\tabcolsep}{3.1pt}
  \renewcommand{\arraystretch}{1.16}
  \resizebox{0.95\textwidth}{!}{%
        \begin{tabular}{
  lc
  cc
  cc
  cc
  cccc
  cc
}
  \toprule
  &
  & \multicolumn{2}{c}{T1}
  & \multicolumn{2}{c}{T2}
  & \multicolumn{2}{c}{T3}
  & \multicolumn{4}{c}{T4}
  & \multicolumn{2}{c}{T5: Q1} \\
  \cmidrule(lr){3-4}
  \cmidrule(lr){5-6}
  \cmidrule(lr){7-8}
  \cmidrule(lr){9-12}
  \cmidrule(lr){13-14}

  Agent
  & \#Params.
  & Acc. & $\kappa$
  & Acc. & $\kappa$
  & Acc. & $\kappa$
  & F1 & MCC & $\rho_{\mathrm{H}}$ & $\tau_{\mathrm{H}}$
  & Acc. & $\kappa$ \\
  \midrule

  Humans
  & --
  & 0.802 & --
  & 0.773 & --
  & 0.637 & --
  & 0.476 & 0.311 & -- & --
  & 0.633 & -- \\

  \midrule
  \multicolumn{14}{c}{\textit{VLMs: 7--12B parameters}} \\
  \midrule

  LLaVA-1.6~\cite{liu2024llavanext}
  & 7B
  & 0.499 & -0.001
  & 0.495 & -0.000
  & 0.333 & -0.008
  & 0.219 & 0.002 & 0.038 & 0.024
  & 0.474 & 0.000 \\

  Qwen2.5-VL~\cite{bai2025qwen25vltechnicalreport}
  & 7B
  & 0.550 & 0.099
  & 0.500 & -0.010
  & 0.356 & 0.028
  & 0.331 & 0.073 & 0.152 & 0.104
  & 0.557 & 0.102 \\

  InternVL3.5~\cite{wang2025internvl3}
  & 8B
  & 0.568 & 0.128
  & 0.620 & 0.205
  & 0.388 & 0.073
  & 0.349 & 0.095 & 0.146 & 0.098
  & 0.570 & 0.162 \\

  Qwen3-VL~\cite{bai2025qwen3}
  & 8B
  & 0.652 & 0.245
  & 0.574 & 0.195
  & 0.401 & 0.109
  & 0.378 & 0.138 & 0.288 & 0.212
  & 0.631$^\dagger$ & 0.205 \\

  Gemma-4~\cite{team2026gemma}
  & 12B
  & 0.634 & 0.199
  & \secondvlm{0.669} & \secondvlm{0.284}
  & 0.501 & 0.240
  & 0.354 & 0.115 & 0.311 & 0.238
  & 0.677 & \secondvlm{0.237} \\

  \midrule
  \multicolumn{14}{c}{\textit{VLMs: 27--38B parameters}} \\
  \midrule

  Qwen3.6~\cite{qwen36plus}
  & 27B
  & 0.700 & 0.336
  & 0.646 & 0.259
  & \secondvlm{0.556} & \secondvlm{0.295}
  & 0.390 & 0.166
  & \secondvlm{0.433} & \secondvlm{0.304}
  & 0.681 & 0.159 \\

  Qwen3.8~\cite{qwen38}
  & 27B
  & \bestvlm{0.770} & \bestvlm{0.461}
  & \bestvlm{0.745} & \bestvlm{0.399}
  & \bestvlm{0.595} & \bestvlm{0.330}
  & \bestvlm{0.443} & \bestvlm{0.232}
  & \bestvlm{0.511} & \bestvlm{0.369}
  & \secondvlm{0.683} & 0.171 \\

  Gemma-4~\cite{team2026gemma}
  & 31B
  & \secondvlm{0.757} & \secondvlm{0.430}
  & 0.654 & 0.280
  & 0.548 & 0.283
  & \secondvlm{0.423} & \secondvlm{0.203}
  & 0.308 & 0.235
  & \bestvlm{0.712} & \bestvlm{0.243} \\

  Qwen2.5-VL~\cite{bai2025qwen25vltechnicalreport}
  & 32B
  & 0.609 & 0.173
  & 0.533 & 0.057
  & 0.371 & 0.062
  & 0.327 & 0.063 & 0.233 & 0.167
  & 0.628$^\dagger$ & 0.194 \\

  Qwen3-VL~\cite{bai2025qwen3}
  & 32B
  & 0.722 & 0.393
  & 0.644 & 0.248
  & 0.485 & 0.209
  & 0.409 & 0.189 & 0.354 & 0.262
  & 0.669 & 0.226 \\

  LLaVA-1.6~\cite{liu2024llavanext}
  & 34B
  & 0.499 & 0.025
  & 0.500 & 0.018
  & 0.336 & 0.014
  & 0.240 & 0.018 & 0.001 & 0.000
  & 0.245 & -0.001 \\

  InternVL3.5~\cite{wang2025internvl3}
  & 38B
  & 0.618 & 0.236
  & 0.577 & 0.163
  & 0.397 & 0.094
  & 0.322 & 0.060 & 0.129 & 0.085
  & 0.647$^\dagger$ & 0.205 \\

  \bottomrule
\end{tabular}%

  }
  \endgroup
\end{table*}

\begin{table*}[t]
  \centering
  \caption{
    Accuracy for humans and for the best-performing VLM, Qwen3.8-27B, in reference matching tasks T1..T3. Results are reported overall, and by scene-change type (geometry changes between reference and candidates, or illumination changes between them). Bold values indicate the best-performing agent in each column.
  }
  \label{tab:e1-e3_accuracy_by_change_type}
  \begingroup
  \small
  \setlength{\tabcolsep}{5.5pt}
  \renewcommand{\arraystretch}{1.15}
    \begin{tabular}{lccccccccc}
    \toprule
    &
    \multicolumn{3}{c}{T1}
    & \multicolumn{3}{c}{T2}
    & \multicolumn{3}{c}{T3} \\
    \cmidrule(lr){2-4}
    \cmidrule(lr){5-7}
    \cmidrule(lr){8-10}

    Agent
    & Overall
    & Illum.
    & Geo.
    & Overall
    & Illum.
    & Geo.
    & Overall
    & Illum.
    & Geo. \\
    \midrule

    Humans
    & \textbf{0.802}
    & 0.773
    & \textbf{0.831}
    & \textbf{0.773}
    & 0.744
    & \textbf{0.801}
    & \textbf{0.637}
    & 0.621
    & \textbf{0.653} \\

    Qwen3.8-27B
    & 0.770
    & \textbf{0.799}
    & 0.741
    & 0.745
    & \textbf{0.768}
    & 0.723
    & 0.595
    & \textbf{0.631}
    & 0.558 \\

    \bottomrule
  \end{tabular}

  \endgroup
\end{table*}

In this section we report the main findings, and include additional results and analyses in the \suppref{supplementary material}.

\subsection{Overall performance}
\label{sec:results-accuracy}

We first evaluate models' performances on the five benchmark tasks, and compare them with the human baseline (\Cref{tab:4_overall_model_performance}).
For tasks T1, T2, T3, and T5:Q1, we report accuracy (proportion of correct responses) across tasks.
To evaluate not only model performance, but also agreement, we compute Cohen's kappa between each model and humans on these tasks.
For T4, which allows multiple candidate selections (\Cref{sec:experiments:experiments}), we report the F1-score and Matthews correlation coefficient (MCC)~\cite{chicco2021mcc}, a pair of measures widely used in binary classification~\cite{chicco2020advantages}. Finally, we compute Spearman's correlation between models and humans to assess agreement in perceptual difficulty in T4.

Overall, larger models (27--38B parameters) tend to outperform smaller ones and exhibit stronger agreement with human judgments. We also observe higher performance among recent Qwen models compared with older ones such as LLaVA. These differences may partly reflect advances in visual encoding, such as the SigLIP 2-based encoder~\cite{tschannen2025siglip} used in Qwen3-VL~\cite{bai2025qwen3}, compared with the CLIP-based encoder of LLaVA-1.6~\cite{radford2021learning, liu2024llavanext}. 

Humans systematically outperform models in matching tasks (T1..T4), and this difference in performance is statistically significant (McNemar's test, $p < 0.005$ in all cases). In the no-reference task (T5), differences between models and humans are reduced: the larger models (except LLaVA-1.6) perform either better --Gemma, and newer models of the Qwen family--, or on par --InternVL3.5 and the rest of the Qwen family-- with humans.

\subsection{Robustness to geometry and illumination}
\label{sec:results:ref-t1}

Beyond overall performance, we assess to what extent VLMs differ from humans in how robust their material matching ability is to changes in geometry and illumination. In matching tasks T1-T3, either geometry or illumination varies between reference and candidates, while the other factor remains fixed (\Cref{sec:experiments}). We thus disaggregate the overall accuracy of humans and the best-performing VLM, Qwen3.8-27B, by change type (\Cref{tab:e1-e3_accuracy_by_change_type}), revealing contrasting trends: Qwen slightly outperforms humans under illumination changes, whereas humans show significantly higher accuracy under geometry changes. These findings hold across all three tasks, suggesting distinct sensitivities to scene configurations, with Qwen struggling more to maintain material constancy under geometric variations.

\begin{figure}[t]
  \centering
   \includegraphics[width=\linewidth]{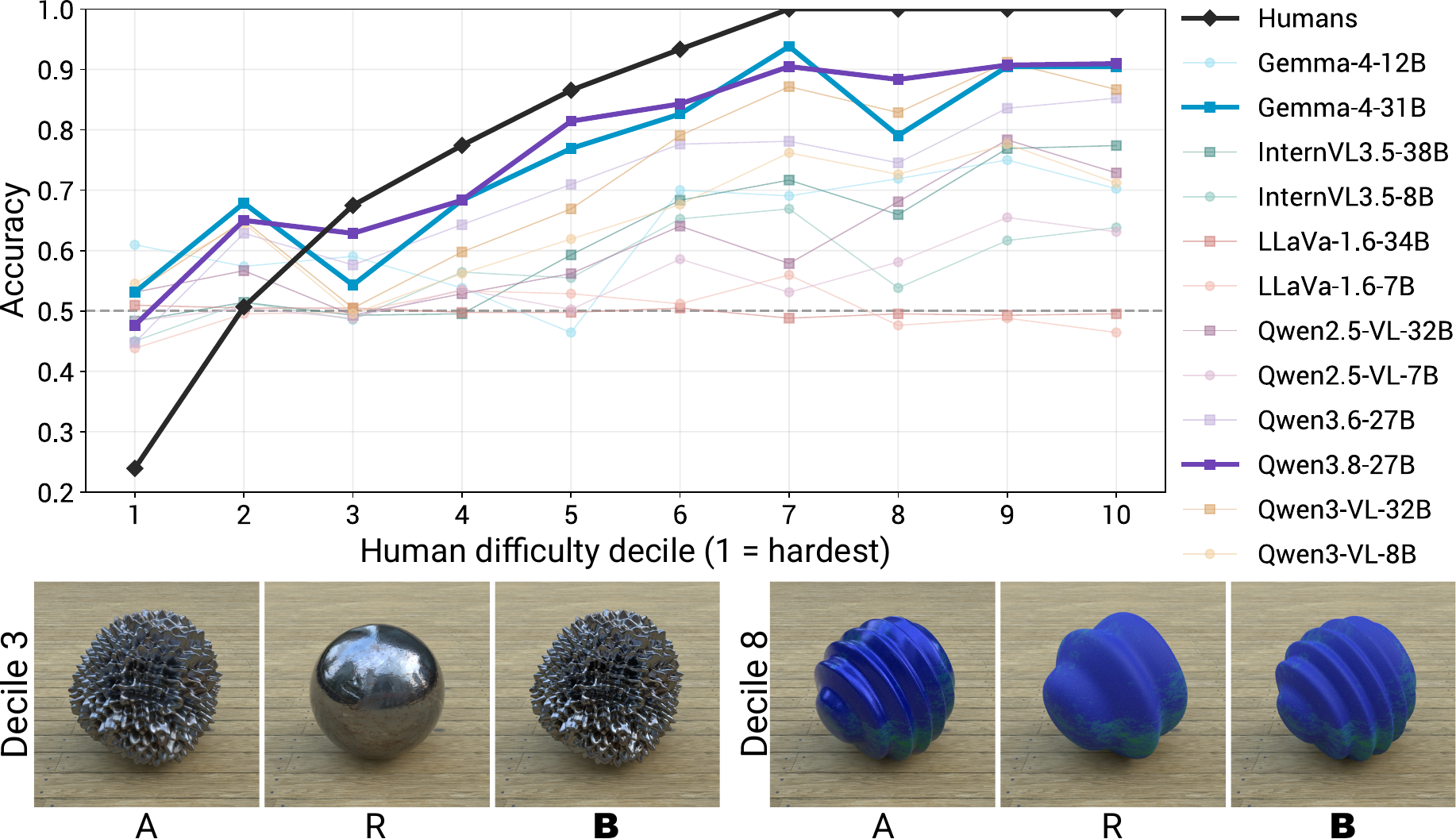}
   \caption{
   \textbf{Top:} Human and VLM agreement on setup difficulty in T1. Setups are ranked by human accuracy and divided into ten difficulty deciles, from hardest (1) to easiest (10). The panel reports per-agent raw accuracy within each decile, emphasizing Humans, Qwen3.8-27B, and Gemma-4-31B.
   \textbf{Bottom:} Representative setups from deciles 3 and 8, with correct answer marked in boldface.
   }
   \label{fig:rq2-difficulty-T1}
\end{figure}

Similar overall accuracy does not necessarily imply similar perceptual behavior, as humans and VLMs may succeed or fail on different stimuli. To investigate whether their perceptual successes and failures align~\cite{storrs2021unsupervised, morimoto2026human, mao2025evaluating}, 
we bin trials into deciles according to observed difficulty for humans, and plot the corresponding accuracy for both humans and VLMs in \Cref{fig:rq2-difficulty-T1} for task T1 (corresponding results for T2 and T3 follow the same trend 
and can be found in the \suppref{supplementary material}). The best-performing VLMs (e.g., Qwen3.8-27B, \Cref{tab:4_overall_model_performance}) are the ones that show higher correlation with humans (Spearman $\rho_{diff}=0.51$),
revealing substantial agreement in perceptual difficulty.
This suggests that some of the same stimuli challenge both types of agents despite their different perceptual capabilities.

\takeawaybox{
    \begin{itemize}
        \item Humans and VLMs differ in their robustness to illumination and geometry changes, with a comparable performance in the former and an advantage of humans in the latter.
        \item Humans and the best-performing VLMs show substantial agreement on which stimuli are perceptually challenging. 
    \end{itemize}
}

\subsection{Influence of context and distractors}
\label{sec:results:ref-t2-t3-t4}

\begin{figure}[t]
  \centering
   \includegraphics[width=0.97\columnwidth]{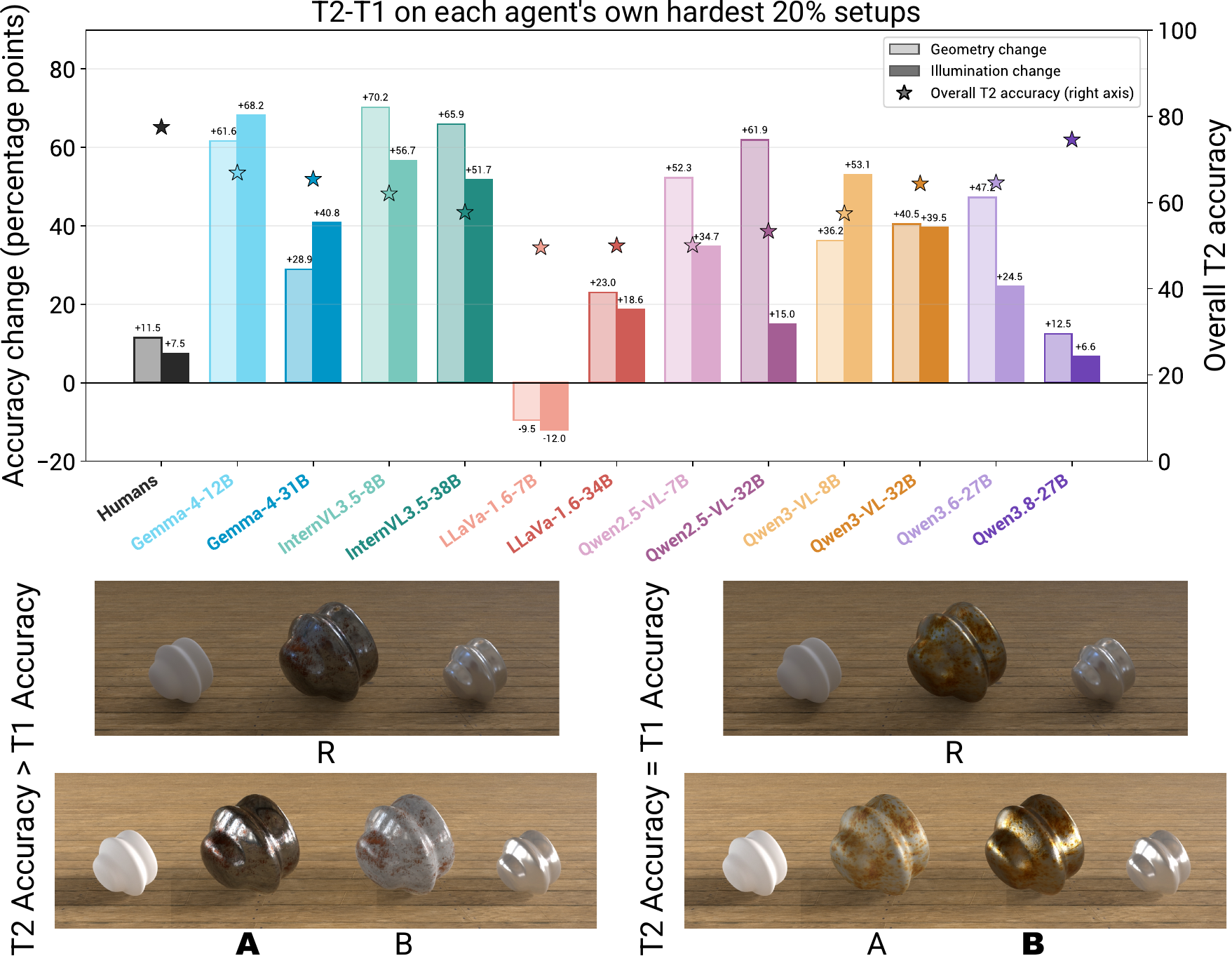}
   \caption{
    \textbf{Top:} T2-T1 accuracy change on each agent's own hardest 20\% of T1 setups, split by geometry (light) and illumination (dark) changes. Stars indicate overall T2 accuracy (right axis).
    \textbf{Bottom:} Examples where context increases accuracy (left) or leaves it unchanged (right).
   }
   \label{fig:rq3-context-hardest-all}
\end{figure}

The design of \bench allows us to isolate the effect of adding visual context by comparing T1 and T2, which share identical reference and candidate stimuli but differ in the presence of contextual probes (\Cref{fig:benchmark_overview}).
While overall performance varies only slightly between T1 and T2, focusing on the hardest 20\% trials reveals a remarkable trend (\Cref{fig:rq3-context-hardest-all}): VLMs are able to successfully leverage the contextual probes to improve their performance. This improvement is more pronounced in smaller models, which exhibit lower baseline accuracies, and is notably greater for geometric than for illumination changes.

T3 extends T2 by adding one more distractor (three candidates) while keeping the contextual probes. We evaluate how this increase in task complexity affects humans and VLMs' performance, and whether that change is consistent over all agents. 
To account for the different chance level,
we use chance-corrected accuracy (CA), defined as $\mathrm{CA}=(\mathrm{Acc}-p_{\mathrm{chance}})/(1-p_{\mathrm{chance}})$, with $p_\mathrm{chance}$ being chance level probability. \Cref{tab:e3-e2_corrected_acc} shows results for the best-performing VLM, Qwen3.8, compared to those of humans. On average, we observe a similar decrease in the chance-corrected accuracy for both agents, experimenting a decrease of 0.085 for the humans, and a decrease of 0.099 for Qwen 3.8. When disaggregating those results by changes in generative factors, we find that Qwen3.8 is more robust to illumination changes, consistent with \Cref{sec:results:ref-t1}, despite exhibiting a proportionally larger performance drop than humans (0.089 vs 0.060). In short, these results reveal performance trend similar to humans' as task complexity is increased by additional distractors.

\begin{table}[t]
  \centering
  \caption{
    Chance-corrected accuracy in T2 and T3 for human observers and Qwen3.8-27B. Results are reported over all setups and disaggregated for geometry-change and illumination-change trials.
    The smaller colored value shown next to T3 is $\Delta=\mathrm{T3}-\mathrm{T2}$ after correcting each task according to their respective chance level.
    }
  \label{tab:e3-e2_corrected_acc}
  \begingroup
  \small
  \setlength{\tabcolsep}{5pt}
  \renewcommand{\arraystretch}{1.14}
    \begin{tabular}{@{}llll@{}}
  \toprule
    & Overall
    & Illum. change
    & Geo. change \\
  \midrule

  \multicolumn{4}{@{}l}{\textbf{Humans}} \\[-1pt]

  T2 corrected acc.
    & 0.550
    & 0.494
    & 0.605 \\

  T3 corrected acc.
    & \scorechange{0.465}{\negative{$-0.085$}}
    & \scorechange{0.435}{\negative{$-0.060$}}
    & \scorechange{0.494}{\negative{$-0.111$}} \\

  \addlinespace[5pt]

  \multicolumn{4}{@{}l}{\textbf{Qwen3.8-27B}} \\[-1pt]

  T2 corrected acc.
    & 0.491
    & 0.536
    & 0.446 \\

  T3 corrected acc.
    & \scorechange{0.392}{\negative{$-0.099$}}
    & \scorechange{0.447}{\negative{$-0.089$}}
    & \scorechange{0.337}{\negative{$-0.109$}} \\

  \bottomrule
\end{tabular}

  \endgroup
\end{table}

\begin{figure}[t]
  \centering
   \includegraphics[width=0.93\linewidth]{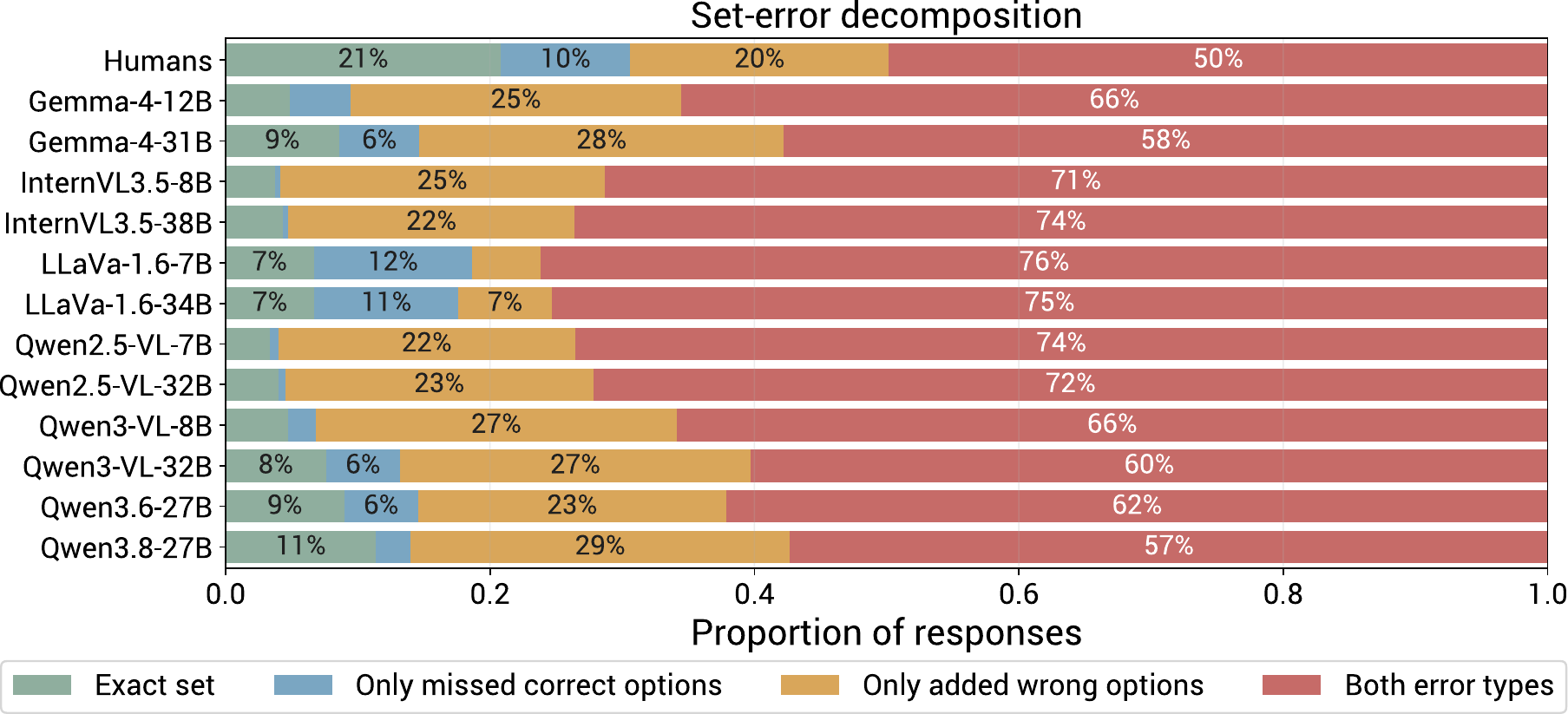}
   \caption{
   Proportion of correct responses and distribution of error types for humans and VLMs. Each bar sums to 100\%.
   }
   \label{fig:4_t4_cardinality_errors}
\end{figure}

Finally, T4 extends T1 by both increasing the number of candidates (from two to six) \emph{and} allowing a variable number of correct matches, introducing the possibility of both false positives and false negatives. Despite similar rates of selecting the correct number of candidates (46--53\%), humans achieve substantially more exact-set matches (21\%) than the best-performing VLM (Qwen3.8-27B, 11\%), highlighting that correct cardinality does not necessarily imply correct material identification (\Cref{fig:4_t4_cardinality_errors}). More importantly, most VLMs exhibit an over-selection bias, choosing too many candidates in 40--50\% of trials, compared to 30\% for humans. Consistently, VLM errors more frequently involve both false positives and false negatives (57--76\%, compared with 50\% for humans). Overall, these findings reveal distinct response strategies, with humans generally adopting a more conservative selection behavior than VLMs.

\takeawaybox{
    \begin{itemize}
        \item VLMs are able to successfully leverage contextual probes to improve their performance on material matching tasks.
        \item VLMs exhibit different response strategies to humans'. Humans are more conservative, leading them to a significantly higher proportion of exact-set responses (2x) than the best-performing VLM. 
    \end{itemize}
}

\subsection{Material identity and perception}
\label{sec:no-reference}

Unlike previous tasks, T5 presents pairs of stimuli without an explicit reference (\Cref{sec:experiments:experiments}), fundamentally changing the nature of the perceptual judgment. Rather than selecting the closest match to a reference, observers now must determine whether two stimuli appear similar enough to represent the same material (Q1). This requires judging \emph{perceptual equivalence} instead of \emph{relative similarity}, involving a different decision process~\cite{garcia2017indecision}. T5 further asks observers to compare perceived glossiness (Q2) and metallicness (Q3), which are subjective judgments for which there is no ground truth. Note that material properties, controlled during stimulus generation, do not uniquely determine perceived attributes, which also depend on scene factors such as geometry and illumination. 
We therefore evaluate VLM behavior using human responses as a baseline. We focus on Qwen3.8-27B
since it is the most aligned with humans in T1..T4, 
and include other models in the \suppref{supplementary}. 

\begin{figure*}[t]
  \centering
   \includegraphics[width=\textwidth]{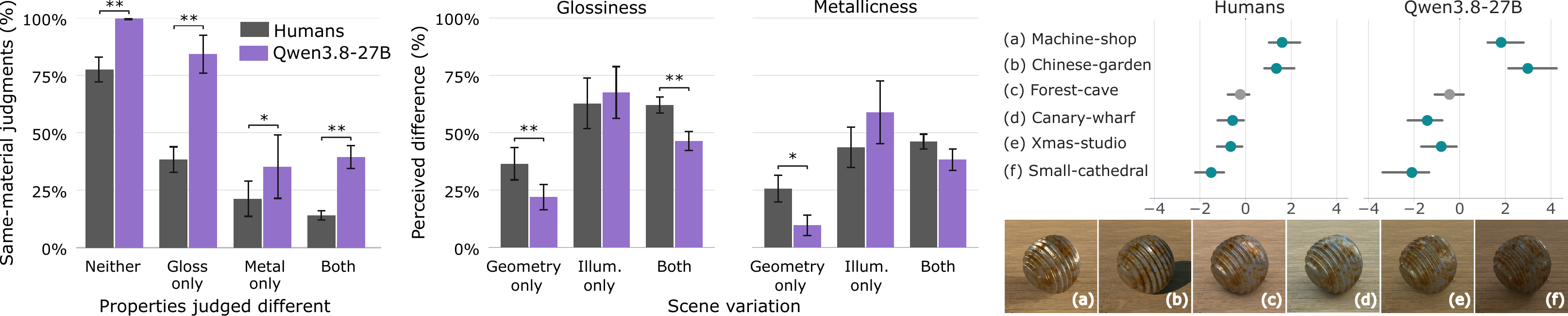}
   \caption{
   \textbf{Left:} Percentage of \emph{same material} responses for stimulus pairs with different underlying material properties, grouped by perceived differences in neither, glossiness only, metallicness only, or both properties.
   \textbf{Center:} Percentage of perceived glossiness (left) and metallicness (right) differences under geometry and/or illumination changes. 
   Error bars show 95\% confidence intervals; asterisks indicate significant differences (paired permutation tests, Holm-corrected).
   \textbf{Right:} Relative glossiness scores across illuminations (from highest to lowest; corresponding illustrative images below). Scores are centered at zero; error bars show 95\% bootstrap confidence intervals. \textcolor{teal}{Teal} markers indicate consistent deviations from the group average.
   }
   \label{fig:4_t5_stats}
\end{figure*}

\paragraph{Appearance perception and material identity.} We first investigate how perceived differences in glossiness and metallicness (Q2 and Q3) relate to material identity judgments (Q1). For this analysis, we consider stimulus pairs in which the material properties (\emph{roughness} and/or \emph{metallic} maps) differ between the two images. We group responses according to whether neither, only glossiness, only metallicness, or both attributes were reported as different between the images, and for each of these four categories, we report the percentage of responses in which the two images were judged to be the same material (\Cref{fig:4_t5_stats}, left). 

Human responses exhibit a clear and consistent pattern: the probability of assessing two images as the same material decreases as more appearance attributes are perceived to differ, following the order of `Neither' $>$ `Gloss only' $>$ `Metal only' $>$ `Both'. Metallicness differences seem to be more strongly associated with material identity judgments than glossiness differences alone~\cite{schmid2023material, chen2021effect}. %

In contrast, Qwen3.8-27B shows a strong tendency to judge materials as identical, reporting the same material more frequently than humans across all conditions, even when the underlying material properties are different, and even when Qwen itself reports differences in both glossiness and metallicness. For instance, when both attributes are perceived as different, Qwen still reports the same material %
in 39.5\% of cases (vs. 14\% for humans). 
These observations suggest that Qwen associates perceived appearance differences with material identity differently from humans.

\paragraph{Scene-dependent material appearance.}
We next measure the proportion of responses reporting an appearance difference in gloss or metallicness for materials with the same underlying material properties (same \emph{roughness} and \emph{metallic} maps) when subject to changes in geometry, illumination, or both (\Cref{fig:4_t5_stats}, center). Even for materials with identical underlying properties, humans frequently perceive appearance differences due to scene variations, consistent with 
previous work~\cite{fleming2017material, lagunas2021joint}. Illumination changes elicit more perceived differences than geometry changes in both glossiness (62.8\% vs. 36.6\%) and metallicness (43.7\% vs. 25.7\%). Qwen shows a similar trend but significantly lower sensitivity to geometry changes than humans (22\% vs. 36.6\% for glossiness; 9.7\% vs. 25.7\% for metallicness), while responding comparably to illumination changes. 

To identify which scene conditions drive these judgments, we fitted an additive Bradley–Terry–Davidson model~\cite{bradley1952rank,davidson1970extending}. This paired-comparison model estimates relative scores from A/B preferences while accounting for ties (\emph{same} responses), allowing us to estimate the contributions of geometry and illumination. \Cref{fig:4_t5_stats} (right) shows that humans and Qwen exhibit remarkably consistent illumination-dependent glossiness trends: \textit{machine\_shop} and \textit{chinese\_garden} yield higher perceived glossiness, whereas \textit{small\_cathedral} yields lower scores. Similar trends are observed for metallicness. 
These findings are consistent with previous research linking high-frequency illumination to increased perceived glossiness-related attributes in humans~\cite{serrano2021materials}. Geometry effects are weaker and less consistently estimated, in line with the lower frequency of geometry-induced appearance differences reported in the previous paragraph. Results for metallicness are provided in \suppref{the supplementary}.

\takeawaybox{
    \begin{itemize}
        \item Human material identity judgments closely reflect perceived appearance differences, while Qwen3.8 shows a systematic tendency to judge materials as identical, even when those differences are apparent to it.
        \item Qwen3.8 captures which illuminations influence material appearance similarly to humans.
    \end{itemize}
}

\section{Discussion}
\label{sec:discussion}

We presented \bench, a benchmark with five tasks designed to investigate material appearance perception 
and constancy 
in VLMs under controlled variations in geometry and illumination, comparing their behavior with that of human observers. 
Our results reveal both similarities and systematic differences, highlighted as insights throughout \Cref{sec:results}.

Our benchmark relies on controlled renderings with a variety of materials, illuminations, and geometries.
While this design is necessary to systematically study the influence of scene factors, it does not capture the full diversity of real-world objects. In particular, the synthetic geometries may partly explain the larger differences between humans and VLMs under geometry changes, and show the higher ability to generalize of humans as compared to VLMs. Future work could complement our experiments with more realistic object geometries or pictures, to see if our observed trend is maintained or whether VLMs catch up. 
Although we focus on gloss and metallicness, two fundamental aspects of surface reflectance~\cite{chadwick2015perception, toscani2020three}, future work could investigate other perceptual properties such as translucency. Our benchmark could also be used to explore proprietary VLMs like GPT and Claude. 
Further investigation of VLMs' internal representations could provide insights into the mechanisms underlying the observed behaviors and inform the development of future models more closely aligned with human visual perception.

Finally, our results advance the evaluation of VLM material perception beyond overall accuracy, revealing where and how their judgments align with or diverge from human perception. By examining both perceptual successes and systematic failures, we provide a more nuanced understanding of human-VLM perceptual alignment, without assuming that behavioral similarities imply shared underlying mechanisms.

\section*{Acknowledgements}

This work has received funding from the European Union (ERC grant number 101220555, PROXIE). Views and opinions expressed are however those of the author(s) only and do not necessarily reflect those of the European Union or the European Research Council Executive Agency. Neither the European Union nor the granting authority can be held responsible for them. 
Additionally, Santiago Jimenez-Navarro was supported by a Gobierno de Aragon predoctoral grant (EMC/590/2025).
This work has also received support from grant PID2022-141539NB-I00, funded by MICIU/AEI/10.13039/501100011033 and by ERDF, EU.
Finally, we want to especially thank Julia Guerrero-Viu for her help designing the final figures and proofreading, and Daniel Jiménez-Navarro for proofreading. We extend our gratitude to the participants of the user study and the members of the Graphics and Imaging Lab.

{
    \small
    \bibliographystyle{ieeenat_fullname}
    \bibliography{main}
}

\end{document}